\documentclass[conference]{IEEEtran}
\IEEEoverridecommandlockouts
\usepackage{cite}
\usepackage{amsmath,amssymb,amsfonts}
\usepackage{algorithmic}
\usepackage{graphicx}
\usepackage{textcomp}
\usepackage{xcolor}
\def\BibTeX{{\rm B\kern-.05em{\sc i\kern-.025em b}\kern-.08em
    T\kern-.1667em\lower.7ex\hbox{E}\kern-.125emX}}
\begin{document}


\title{Olfactory Inertial Odometry: Methodology for Effective Robot Navigation by Scent}

\author{\IEEEauthorblockN{Kordel K. France}
\IEEEauthorblockA{\textit{Dept. of Computer Science} \\
\textit{University of Texas at Dallas}\\
Richardson, TX, USA \\
kordel.france@utdallas.edu}
\and
\IEEEauthorblockN{Ovidiu Daescu}
\IEEEauthorblockA{\textit{Dept. of Computer Science} \\
\textit{University of Texas at Dallas}\\
Richardson, TX, USA \\
ovidiu.daescu@utdallas.edu}
}

\pagestyle{plain}

\maketitle

\begin{abstract}
Olfactory navigation is one of the most primitive mechanisms of exploration used by organisms. Navigation by machine olfaction (artificial smell) is a very difficult task to both simulate and solve. With this work, we define olfactory inertial odometry (OIO), a framework for using inertial kinematics, and fast-sampling olfaction sensors to enable navigation by scent analogous to visual inertial odometry (VIO). We establish how principles from SLAM and VIO can be extrapolated to olfaction to enable real-world robotic tasks. We demonstrate OIO with three different odour localization algorithms on a real 5-DoF robot arm over an odour-tracking scenario that resembles real applications in agriculture and food quality control. Our results indicate success in establishing a baseline framework for OIO from which other research in olfactory navigation can build, and we note performance enhancements that can be made to address more complex tasks in the future.
\end{abstract}

\section{Introduction}
From the first life forms to complex mammals, the ability to navigate using scent has been a cornerstone of survival. 
Animals like ants, hounds, and rodents demonstrate remarkable proficiency in following odour plumes and pheromone trails to locate food, mates, or shelter. These feats are achieved through a sophisticated interplay between acute scent receptors and motion. 
However, the physical behavior of odour plumes—constantly shifting with wind, influenced by temperature and humidity, and weakening over time—presents a formidable challenge. When the odour source is out of sight, organisms rely entirely on olfactory cues, transforming the task into a complex control problem that demands robust uncertainty management.

\begin{figure}
  \centering
  \includegraphics[width=85mm]{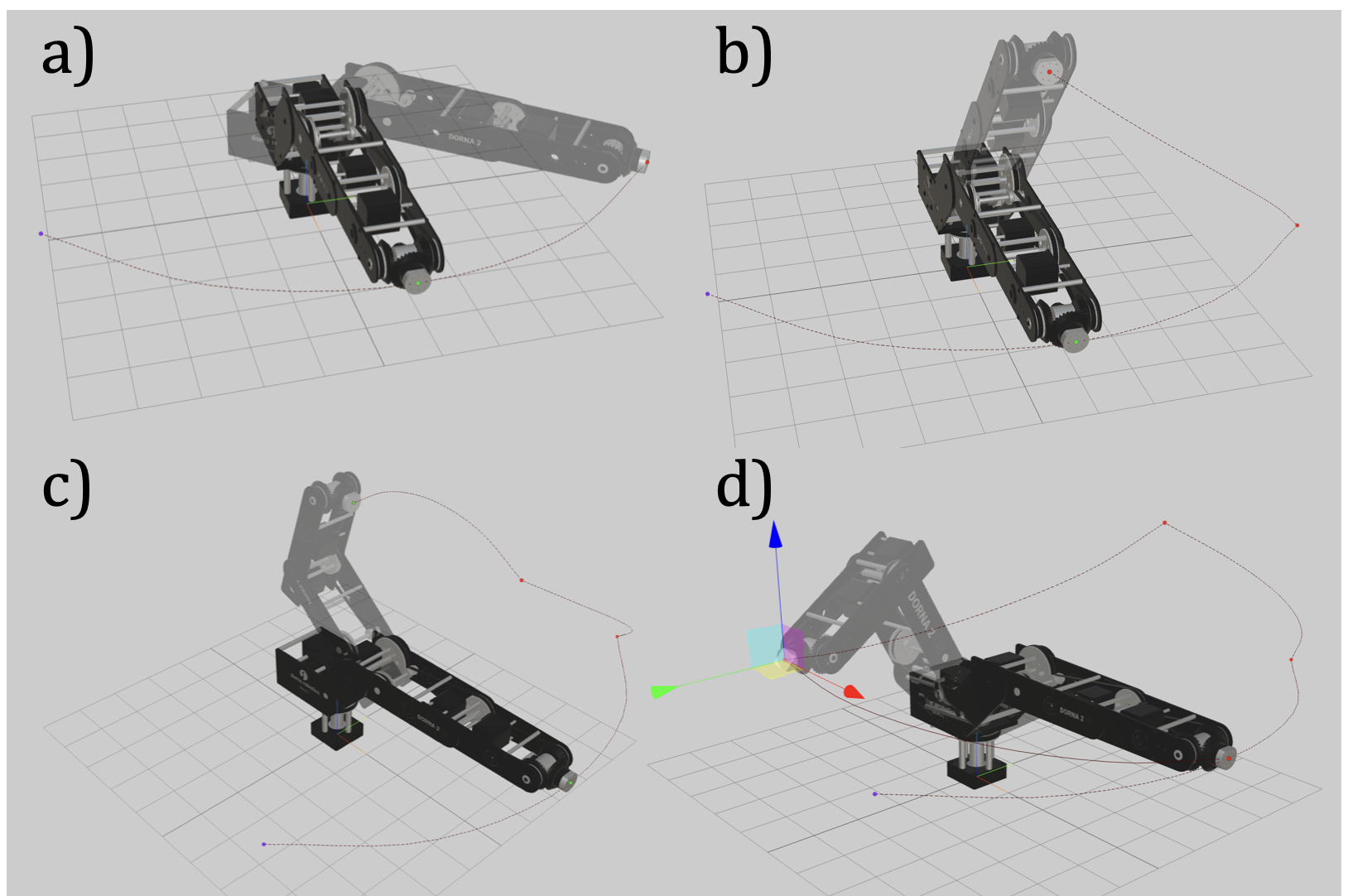}
  \caption{Illustration of the degrees of freedom enabled by the robot arm. a) 1-DoF: Only azimuth in the shoulder. b) 2-DoF: Azimuth and elevation in the shoulder. c) 3-DoF: Azimuth and elevation in the shoulder plus tilt in the elbow. d) 5-DoF: Azimuth and elevation in the shoulder, tilt at the elbow, and both tilt and roll at the wrist.}
  \label{fig:robotdof}
\end{figure}

In robotics, machine olfaction—the artificial sensing and processing of odours—has the potential to revolutionize applications like robot-assisted medical surgery, hazardous material detection, and food and cosmetic quality control. 
Yet, olfactory navigation remains under-explored compared to vision and auditory-based systems, largely due to the unpredictable nature of odour dynamics and the difficulty of integrating scent data with precise motion control. 
In this work, we introduce Olfactory Inertial Odometry (OIO), a novel framework inspired by visual-inertial odometry (VIO) techniques, to address this gap and enable robots to track and locate odour sources with greater precision.
We intend our research to contribute to more applications in robot-assisted surgery and quality control in chemical manufacturing, such as perfumery.

Our approach combines olfactory sensors, inertial odometry, and simultaneous localization and mapping (SLAM) to guide a robotic arm in tracing a chemical plume back to its source.
Effectively, we replicate a scenario where a pick-and-place robot may need to distinguish between multiple chemical samples in order to determine which one contains a target compound.
Vision is typically reserved for pick-and-place tasks, but scenarios involving chemical quality control exceed the limits of vision sensors.

Our approach draws inspiration from biology and explores parallels between robotic and animal navigation, leveraging concepts like gradient optimization and reinforcement learning to mimic scent-tracking behaviors.
We hypothesize that foundational gradient-based optimization techniques can effectively navigate chemical plumes but propose that reinforcement learning (RL) offers an additional edge in dynamic, time-sensitive tasks. 
By directly integrating olfactory signals into the reward function, RL enables adaptive plume-tracking strategies that balance speed and accuracy. 
Furthermore, we demonstrate these techniques over two olfactory sensor types: electrochemical and metal oxide sensors.

Ultimately, our work demonstrates that OIO is a powerful principle for olfactory navigation in pick-and-place tasks, with implications far beyond static or terrestrial robots. 
While this study focuses on low-DoF applications, the methodology and insights gained lay the groundwork for future exploration in high-DoF systems like robotic aerial vehicles. 
As the odour gradient naturally decays and equalizes with air over time, rapid and accurate plume tracking becomes critical—a challenge our framework is designed to address.

Our contributions can be summarized as follows: 
\begin{enumerate}
\item We illustrate how visual inertial odometry techniques can be intuitively extrapolated to support olfaction by fusing olfactory data with inertial data.
\item We demonstrate a framework called OIO for precisely localizing an odour source over two different types of olfaction sensors to show generalization.
\item We evaluate the methodology of OIO over three different tracking algorithms with a real robotic arm localizing itself to an odour source.
\end{enumerate}

\section{Background}
Real-world application of olfactory pick-and-place reside in crop health monitoring in agriculture \cite{chen-plant-detection-2024}, food spoilage detection, and cosmetic quality control.
VIO and SLAM navigation techniques enable a robotic agent to navigate the world around it through visual perception and motion.
Inertial measurements allow the robot to measure the temporal difference between visual samples (e.g. camera frames) and interpolate how it has moved about the environment between time steps. 
Visual processing is very effective, but can be high bandwidth and ineffective in low-light, smoky, or other environmentally-obscuring conditions.
Animals, such as hounds, use VIO principles coupled with their highly sensitive olfactory bulb to detect narcotics and triangulate the source of explosives.
Thus, incorporating another modality such as scent into robotic navigation is not only compelling, but proven to be of real-world utility.
Some research suggest that olfaction is a spatial sense not unlike vision which creates a convenient analog between VIO and OIO.

One common pain point among machine olfaction research used to be the long sampling times needed for a chemical sensor to detect (or react) with the target compound. 
However, recent advancements in olfaction technology have enabled multichannel chemical sensors to respond to stimuli on the order of seconds.
Work from \cite{dennler24} shows how metal oxide sensors can be sampled several times per second through clever electrical design.
We leverage a regression technique to extend a similar response time to electrochemical sensors and effectively evaluate two sensor types over the same experiments.
This decrease in sampling time allows us to leverage multiple techniques to counteract hysterisis and enables the use of real-time sensor fusion algorithms.

The focus of this work is to illustrate a methodology for olfactory-based navigation and some optimizations that make the idea more tractable.
We are specifically interested in assessing the efficacy of olfactory inertial odometry at the edge for real-time robotics applications.
With our robotic arm, we have access to inertial kinematics in each joint which allow us to fuse inertial odometry techniques with the olfactory signal, analogous to how vision and inertial data are fused for VIO.
We leverage gradient optimization algorithms to illustrate that our methodology is inherently enabled by the optimization of sensor and processes levels, rather than an exotic deep learning or neural network architecture.
We build off the work of \cite{dennler24} in detection methodology and general principles for experimental setup.

\section{Related Work}
\label{sec:related}
With OIO, our intention is to investigate a derivative of Burgués's 2019 work \cite{sniffybug-single-burgues19} in odour tracking but with relaxed constraints, a different set of algorithms, and a robotic arm contextualized for pick-and-place and tactile sensing applications.
Our analysis of related work contains references to algorithms that ultimately act as the building blocks of OIO, but it is to our knowledge that no prior work exists surrounding the exact methodology proposed in this paper, nor have we observed this methodology being replicated on an actual robot.

\subsection{Inertial Odometry}
Odometry is a technique that uses motion sensor data to estimate the position of a robot over time.
The motion data can come from several different elements such as translational or rotational movments from wheels, gears, magnets, tracks, or other motion-driven effectors.
Visual odometry extends general odometry by adding perception sensors, such as cameras or lidar from which motion data is interpolated.
It is a process for determining a robot's current location by analyzing the changes in location of points of reference between different camera frames.
When coupled with inertial measurement units, visual inertial odometry can be an accurate means of navigation through the fusion of both visual frames of reference and changes in kinematics.
Dead reckoning is the process of deducing one's current position via analysis of previous positions and current estimates of velocity, acceleration, and other kinematic elements.
OIO is a confluence of inertial odometry and dead reckoning, and we show how this confluence is enhanced by reinforcement learning.

\subsection{Machine Olfaction}
Machine olfaction is a science for which several sensing mediums exist, but with little training data.
This makes olfactory applications particularly interesting because the data intensive methods from other machine learning applications do not apply.
The research of Jas Brooks \cite{jasbrooks2021} covers several machine olfaction applications, but does not have a heavy focus on robotics.
Burgues, et al. in \cite{burgues20-drone-chem-sense-survey} cover many current olfactory applications in robotics, but none seem to illustrate multiple olfaction sensors navigating on the same platform.
\cite{sniffybug-single-burgues19} demonstrate navigation to a gas source with a single UAV.
Duisterhof, et al. extend this work in \cite{sniffybug-swarm-duisterhof21} by demonstrating a swarm of the same UAVs localizing an odour source.
Both emphasize compelling results, but they primarily focus on the use of metal oxide sensors as the detectors and do not seem to get within less than a half-meter of the odour source.
We attempt to demonstrate how OIO can approximate an odour source to within centimeters and build much off of their work.

\cite{Singh2023} show how emergent behavior develops in odour tracking swarms and emphasize the use of RL as a contributor to successful plume tracking.
The work of \cite{France2024} build on their RL model, but instead show how building a policy to seek the expected reward over one that seeks the maximum reward is more robust to aleatoric uncertainty present in odour plumes.
However, their experiments focus on simulation and never enter the real world.
We leverage their reward model for OIO and demonstrate its effectiveness in applied olfactory navigation for short-distance tasks that mimic, for example, robotic surgery, chemical quality control on a manufacturing line, and food spoilage monitoring.

\section{Methodology}
With our robot, we have access to high-order kinematics through the firmware.
We measure angular position (as determined by each joint motor's encoder), velocity, acceleration, and jerk of each joint.
The coordinates for the effector (or "fingertip") at the end of the final joint are located in 3-dimensional Cartesian space (\textit{x-position, y-position} and \textit{z-position} with 2 dimensions of pose (\textit{azimuth} and \textit{elevation}.
Commands are given to the robot to optimize the position of the effector so that it moves as close to the plume source as possible.
We run our experiments with three different optimization methods: a basic closed-loop feedback algorithm, a radio-frequency inspired belief-map algorithm, and one reinforcement learning algorithm.
\subsection{Sensing Techniques}

Of the many olfactory sensing techniques that exist, metal oxide sensors are among the fastest and most sensitive. 
They have been widely used in olfactory navigation tasks \cite{sniffybug-single-burgues19} \cite{sniffybug-swarm-duisterhof21} and their resilience in various environments has been demonstrated by \cite{dennler24} \cite{Schmuker} \cite{burgues20-drone-chem-sense-survey}. 

We select a pair of Sensirion MICS6814 metal oxide (MOX) sensors for our first sensor variant.
These MOX sensors are sensitive to a variety of gases, but we focus on their sensitivity to Benzene for our purposes here.
The MICS6814 changes resistance based on its exposure to the target gas.
This resistance is an indication of gas intensity in the surrounding environment and can be computed according to the following equation:

\begin{equation}
    R_s = \left [ \frac{V_c}{V_{RL}} - 1\right ] \cdot R_L
    \label{eq:mox}
\end{equation}

\noindent 
where $R_S$ indicates the sensor resistance, $V_c$ is the circuit voltage, $V_{RL}$ is the voltage drop over the load resistor, and $R_L$ is the resistance of said load resistor.
We build off of the methodology from Burgués, et al. in \cite{sniffybug-single-burgues19} in that we design our algorithm to not be dependent on absolute concentrations in part-per-million or part-per-billion quantity.
Rather, we analyze the relative change between timesteps of $V_{RL}$ and use this as our metric for gas measurement with the metal oxide sensors.

We have a desire to show how OIO can be leveraged among olfaction in general and not specific to one sensing technique, so we select electrochemistry as the second sensing type for which we demonstrate our techniques.
There are several different sensing techniques in electrochemistry and we select chronoamperometry due its fast sampling ability and temporal alignment.
Chronoamperometry is the process by which the change in electric current with respect to time over a controlled electrical potential can be measured.
Its success in olfaction has been demonstrated by \cite{GraefPrasad2017} \cite{PaulPrasad2020} and \cite{Banga2022}.
Chronoamperometry is governed by the Cottrell equation: 

\begin{equation}
    I = \frac{n_eFAc_k \sqrt{D_k}}{\sqrt{\pi t}}
    \label{eq:cottrell}
\end{equation}

\noindent where $I$ denotes the electrical current, measured in amperes; $n_e$ is the number of electrons needed to oxidize one molecule of analyte $k$; $F$ is the Faraday constant of 96,485 Coulombs per mol; $A$ denotes the planar area of the electrode in square centimeters; $c_k$ defines the initial concentration of the target analyte $k$ in mols per cubic centimeter; $D_k$ defines the diffusion coefficient for analyte $k$ in square centimeters per second; and $t$ is simply the time the chronoamperometric sequence is running in seconds. For the purposes of our sensor which contains an electrode surface area of 2.25 square centimeters and is specifically tracking a target analyte with a reduction potential of 0.8 V according to Fick's law.

As a means of control, our agent is guaranteed to start each episode of training within the simulation outside of the plume in order to get an appropriate baseline measurement.

Olfactory signals are highly influenced by aerodynamic factors such as temperature, pressure, and relative humidity.
Temperature and humidity sensors are attached next to each of the olfaction sensors on the robot, so both variables are measured each time an olfaction sensor is sampled.
The data from both the metal oxide and electrochemical sensors are corrected in accordance to the observed temperature and humidity.
Prior experimentation found that only large changes in environmental pressure influence the sensor responses, so we do not include a pressure sensor on the robot tool.
With all of these sensors, the sensing tool is computing the concentration of the target analyte present in the surrounding air in order to determine its proximity to the much larger plume being emitted to the source. 
Stronger concentrations of the target analyte indicate that the agent is moving closer to the plume source. 
Each action the robot takes in the environment considers a sensing output conditioned on the above that takes 1.0 second to measure. 

\begin{figure}
  \centering
  \includegraphics[width=85mm]{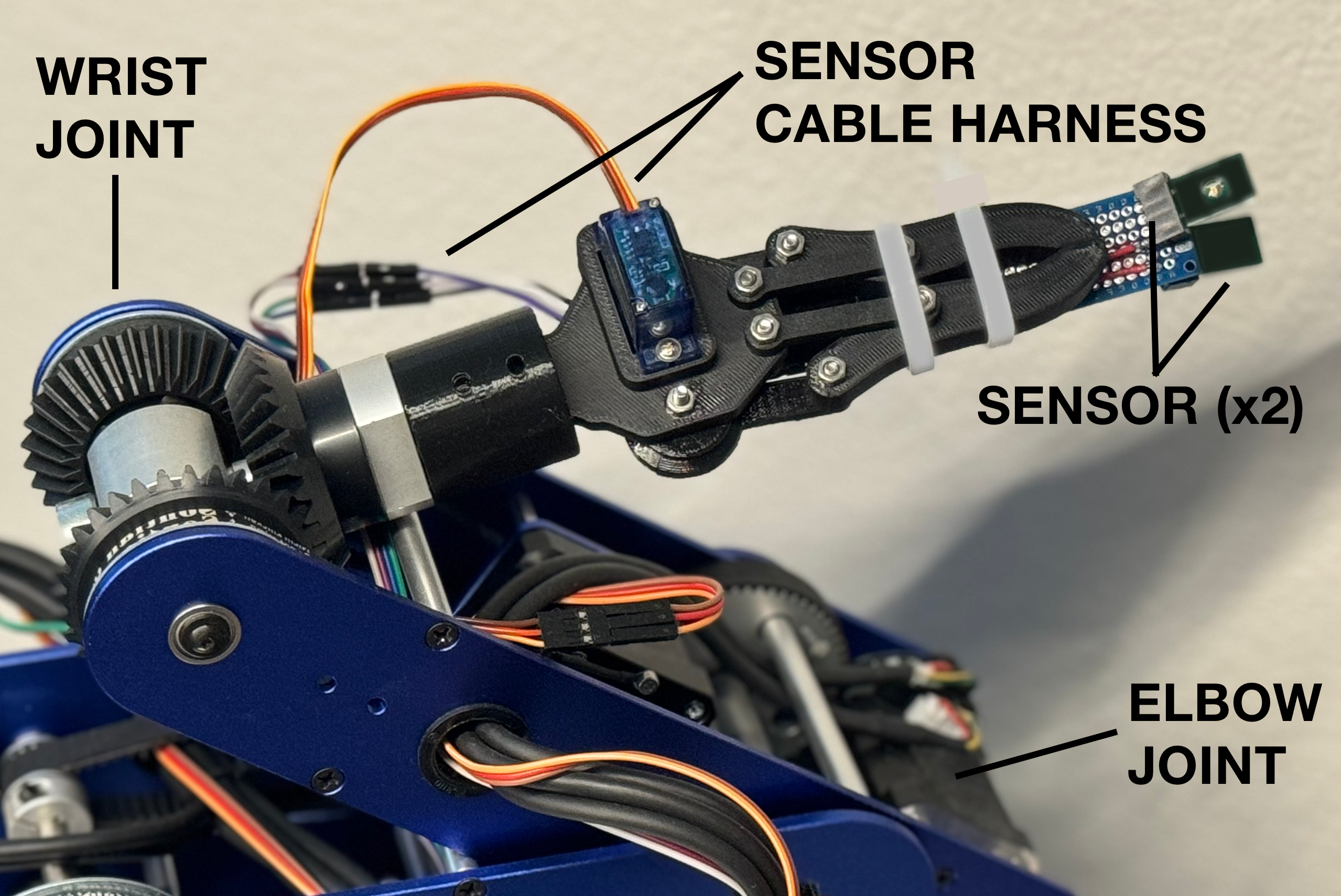}
  \caption{A diagram showing how the electrochemical sensors attaches to the robot "finger tip". The metal oxide sensors attach in a similar manner.}
  \label{fig:robotTool}
\end{figure}

For redundancy in sampling, we leverage two sensors for "sniffing" the air that are integrated into the "finger tip" of the robotic arm.
One sensor is enabled, while the other is disabled, and this oscillatory process repeats the entire time that the robot scans for a signal.
Both sensors are located on the "finger tip" of the robotic arm, and a diagram of this is shown in Figure \ref{fig:robotTool}.
We place each sensor on a 100mm extending board to maximize passive airflow.
The signals between both of these sensors are simply averaged here, but a future work will investigate how the disparity between both sensors can inform the direction in which the robot will move - much like stereo smell in \cite{jasbrooks2021}.

Plumes of air growing from a target source typically have the highest concentrations of the target analyte near the source.
With time, the plume expands and edges of the plume are of lowest concentration.
One might recognize that this can enable simple gradient optimization as an agent can simply follow the gradient of the plume continuously toward increasing concentration until it reaches the plume source.
However, due to wind shifts and environmental factors that influence aerodynamics of ambient air, the gradient to optimize becomes much more nuanced and therefore complicates the search space.
This led us to evaluate RL as part of OIO to determine if a more advanced machine learning algorithm could improve plume tracking over simple gradient-following counterparts.

\vspace{-2mm}
\subsection{Landmark Estimation and Fusion}
We design our experiments to replicate a "pick-and-place" task commonly seen with robotic arms.
The pick-and-place task replicates the principles of agricultural health monitoring \cite{chen-plant-detection-2024} or food spoilage \cite{wang2025-pork-freshness}, and robot-assisted surgery.
For each simulation, the robot is programmed to locate the source of the plume.
The robot may predict one or many plume source locations, and we call each of these locations "landmarks".
Landmarks are filtered by their level of uncertainty so that the robot only navigates the effector to the most probable landmark.
In a real application, a landmark would be the vial or specimen of interest for comparison to a baseline.
As the robot moves, it may adjust the order of landmarks in its bank by re-assessing their uncertainties as a function of their changing covariances.
To establish these landmarks, the robot uses an extended Kalman filter (EKF).
Extended Kalman filters are an adaptation of regular Kalman filters for non-linear problems.



Both the process and observation noise models for the EKF are assumed to be zero-mean multivariate Gaussian distributed with known covariances.
Prior experimental calibration for each sensor type established the covariances for each sensor type up to 30 minutes of use \footnote{Olfaction sensors are known to exhibit drift and non-stationarity during long continuous runtimes. While our experiments verify the noise distributions for up to 30 minutes of use, we suspect an increase in stationarity beyond this.}.

\begin{figure}
  \centering
  \includegraphics[width=85mm]{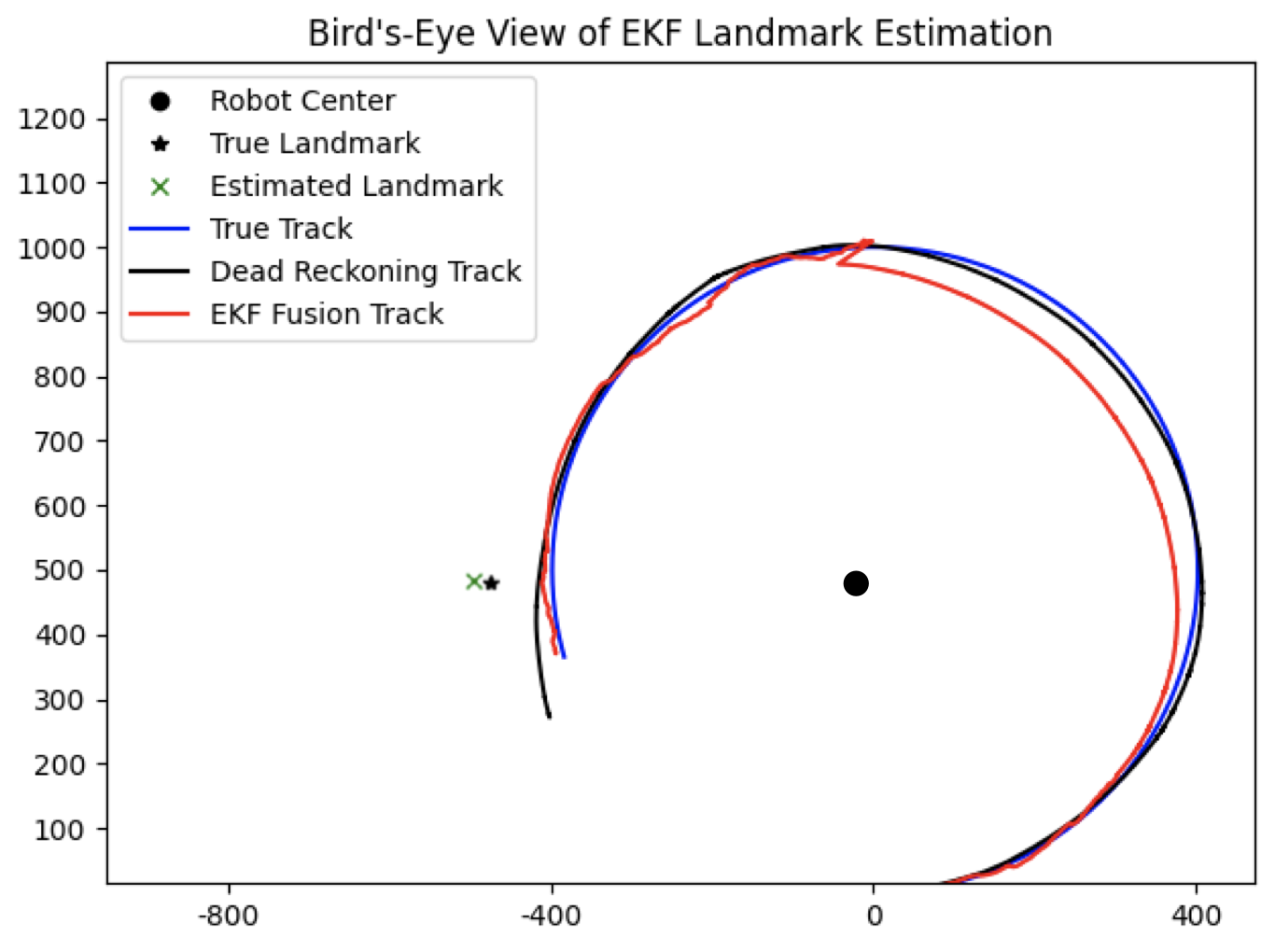}
  \caption{A diagram of the EKF state estimates compared to the true track and the dead reckoning track. The EKF landmark estimate of the odour source is also shown next to the true odour source position.}
  \label{fig:oio-ekf}
\end{figure}

The encoder drift for each limb in the robot is denoted in Table \ref{tab:table1}, and we leverage these values when tuning the EKF.

\vspace{-1mm}
\begin{table}[htbp]
\caption{Encoder drift per limb}
\begin{center}
\begin{tabular}{|c|c|}
\hline
\textbf{Limb}& \textbf{deg/sec2} \\
\hline
Base& 2e-2\\
Shoulder& 1e-2\\
Elbow& 3e-3\\
Wrist (tilt)& 1e-3\\
Wrist (rotation)& 1e-3\\
\hline
\end{tabular}
\label{tab:table1}
\end{center}
\end{table}

\subsection{Bout Detection}
The beginning of each experiment occurs \textit{after} the warm-up period has finished for each sensor type.
For metal oxide sensors, this is on the order of 1 hour and for electrochemical sensors, this is on the order of 1-2 minutes of continuous sampling.
After this, five initial measurements are taken $[y_0, ... , y_4]$.

Schmuker, et al. \cite{schmuker16-bout-detection} define a method for bout detection in olfactory signals.
They establish their method for metal oxide sensors, but we leverage the same principles for the electrochemical sensors with some refinements suggested by \cite{dennler22drift}.

To filter the olfactory signal and provide smoother gradient following, we define a window of length 5 around each of the raw measurements $y_t$ and compute the moving average $y'_t$ at each time step.
Each measurement $y'$ at time $t-1$ is subtracted from the measurement $y_t$ at the current time $t$, effectively measuring the differential of $y_t$: 

\begin{equation}
    \delta_t = y'_t - y'_{t-1}
    \label{eq:bout}
\end{equation}

If the differential $\delta_t$ exceeds the previous value $\delta_{t-1}$ \textit{and} the current absolute value $y'_t$ is also greater than the initial baseline measurements taken at the beginning of the experiment $[y_0, ... , y_4]$, we assume the robot is moving toward the source of the odour. 
While the moving average window was tuned experimentally, we suspect that this value will be dependent on the environment in the future and thus have to be automatically calibrated.

We leverage the above method to provide the output for each sensor for all three experiments.
\subsection{RSSI-Inspired Belief Maps}
In signal processing, the received signal strength indicator (RSSI) is a common path-loss value used to interpret how close a receiver is to a transmitter.
We extend some of these practices from conventional signal processing in order to turn the olfactory sensor response into a RSSI value. 
As each olfaction sensors gets closer to the target compound, its value gets larger.
With the metal oxide sensors, this value is a resistance ratio; with the electrochemical sensors, this value is an amperometric ratio.
RSSI values are typically measured in decibel-milliwatts (dBm), where a higher negative value indicates a weaker signal.
To convert the olfaction signal into this negative format, we simply invert the olfaction values and negate them, which allows us to maintain the same conventions of a lower negative value being analogous to a higher signal strength (and thus closer to the target compound).

As these olfaction RSSI values are received from the sensor, they create a belief map of where the plume is located conditioned on the internal kinematics of the robot arm.
Fundamentally, each RSSI value represents the radius of a sphere that contains equally probabilistic points on that sphere's surface of where the target compound could be located.
We call these points "sigma points" in relation to those generated by unscented Kalman filters.
In other words, if one imagines that the sensor is the center of a sphere, and the sensors' averaged response represents the sphere's radius, the target compound should have an equal probability of lying anywhere on that sphere's surface.
While this narrows down the possible locations of where the target compound could be, there are infinitely many points that lie on the surface area of the sphere.
By moving the sensor through a robot command, we can re-sample the sensors at a different location and obtain a second sphere of probabilities.
These two spheres should intersect, and the intersection of two spheres is a circle.
So now, the possible locations of where the target compound lies is located on that circle's circumference, and this circumference forms the new series of sigma points.
By moving again, we obtain a third sphere, ideally intersecting the former two spheres.
If the intersection of two spheres forms a circle, then the intersection of three spheres should form two circles that will themselves intersect at two points.
This significantly narrows the probability space of where the plume source is located down to only a pair of sigma points.
From here, the robot randomly selects one of the two points to move towards. 
If the RSSI gets larger, then the robot is likely moving toward the right point and should continue to do so until it reaches the target compound and/or reaches its envelope of movement.
If the RSSI gets smaller, then the robot is likely moving away from the plume source and should change directions toward the other point.
This last movement creates a fourth probabilistic sphere.
If all four of these spheres intersect, they will do so at a singular point which will theoretically be the location of the plume source.
The robot should then continue to move toward this point for the duration of its navigation.
In geometry, this is known as the Voronoi vertex.

The above describes the ideal scenario for eliminating uncertainty and constructing a belief map of possible plume source locations through trilateration.
In a perfect world, there would only be four movements needed to exactly locate the source of the target compound. 
In reality, there is a high degree of stochasticity that prevents things from working out quite so simply.
Plumes are extremely dynamic and can shift with time so there is likely to be a lot of backtracking and much more than four movements.
Even the robot movement itself can influence plume changes and induce more variability into navigation.
In addition, there is drift from the olfaction sensors and tolerances in the motor encoder positions of the robot that must be considered.
To aid with this, we recursively create spherical belief maps from the robot's movement and only analyze the last five spheres.
This allows the robot to "forget" older points of belief that may no longer be associated with the current plume due to its diffusion.
Additionally, since there are two sensors performing simultaneous measurements, we create an uncertainty window around the sphere such that the larger value forms the upper bound of the window, the smaller value forms the lower bound of the window, and the average of the two values forms the general radius. We denote this method as the "belief map method" in our experiments.

\begin{figure}
  \centering
  \includegraphics[width=85mm]{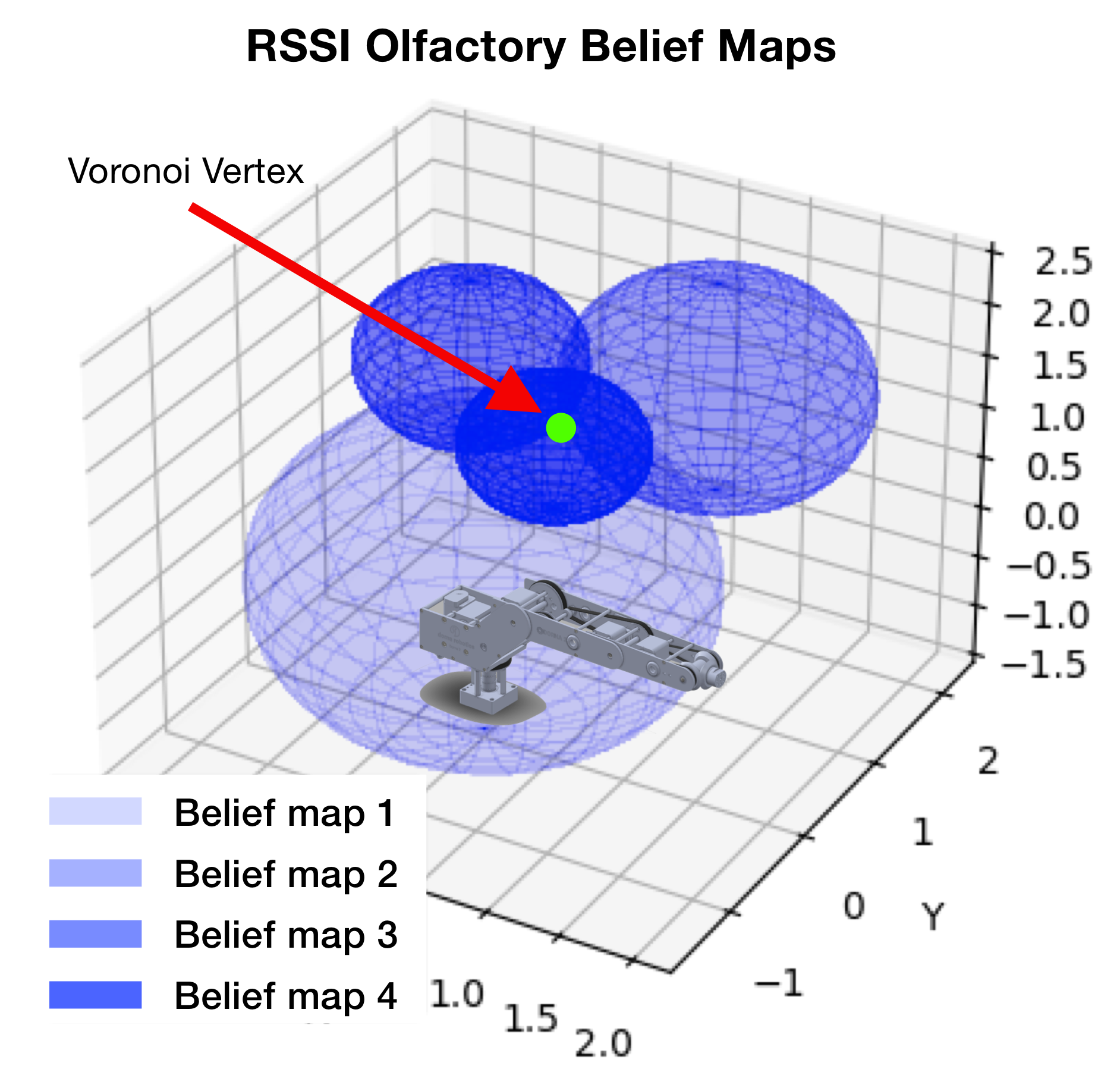}
  \caption{Four belief map spheres intersecting to form the Voronoi vertex. While only four spheres are theoretically needed to find the Voronoi vertex, the robot shows many more are needed during experiments.}
  \label{fig:rssi-belief-maps}
\end{figure}

\subsection{Modeling the Olfactory Reward Signal}
Reinforcement learning typically models a Markov decision process (MDP) characterized by the tuple $(S, A, \mathcal{T}, R, \gamma)$. The state- and action-spaces are represented by $S, A$, respectively. 

The action space of our RL agent enables the robotic arm in our simulation to select an acceleration (in degrees per second) for one second that moves the effector in one of six directions: left, right, up, down, forward, and backward.
We note that an acceleration of zero (remain still) is a possible option.

Singh, et al. demonstrates successful olfactory tracking through an RL model in his 2023 research \cite{Singh2023}. 
We similarly develop a simulation from which we train and calibrate our RL model and build upon many of their assumptions.
The state and action spaces of the robot are not astronomically large.
In addition, both are well-bounded and discretized - the action space is discretized to allow the robot time to sample in between each movement and the state space is discretized such that each position is represented by a single 3-dimensional point.
We strive to demonstrate that the core algorithmic enablement of olfactory tracking comes from the OIO algorithm itself, with RL acting as a secondary complementary layer.

\subsection{Expected Rewards vs Maximum Rewards}
Plume signals are inherently noisy due to their high dimensionality.
Many RL algorithms attempt to maximize the cumulative reward for a task.
While this can be successful in some tasks, we find that support from \cite{Singh2023} \cite{Crimaldi2022DynamicSensing} and \cite{France2024} suggest that plume dynamics are so volatile that the reward signal can be misleading, especially when one accounts for the strong hysterisis present in machine olfaction sensors.
In most state-of-the-art reinforcement learning research, where the environment can be directly observed, Q-learning and its variants balance performance and computational cost compared to other TD algorithms. 
Q-learning is a greedy off-policy algorithm seeking the action that delivers the highest reward.
However, based on the evaluation of \cite{France2023}, Q-learner performance depends on the confidence in observing true rewards for actions. If there is high confidence that the observed rewards match the ground truth, a greedy Q-learner is expected to perform better. 
Therefore, France et al. hypothesize that Expected SARSA may outperform Q-learning in swarm learning, where the Expected SARSA update rule for the next state and action $Q^*(s,a)$ is as follows.

\begin{equation}
     Q(s, a) + \alpha \left [R(s, a, s') +  \frac{\gamma}{n}\sum_{i=1}^{n} Q(s'_i, a_i') - Q(s, a) \right ]
    \label{eq:expectedsarsa}
\end{equation}



Particularly at the start of plume tracking, when the olfactory signal can be erratic due to low concentrations and unstable air, Expected SARSA provides smoother learning. 
Expected SARSA allows our agent to alternate between an on-policy and an off-policy method while naturally controlling reward loss. 

\begin{figure}
  \centering
  \includegraphics[width=85mm]{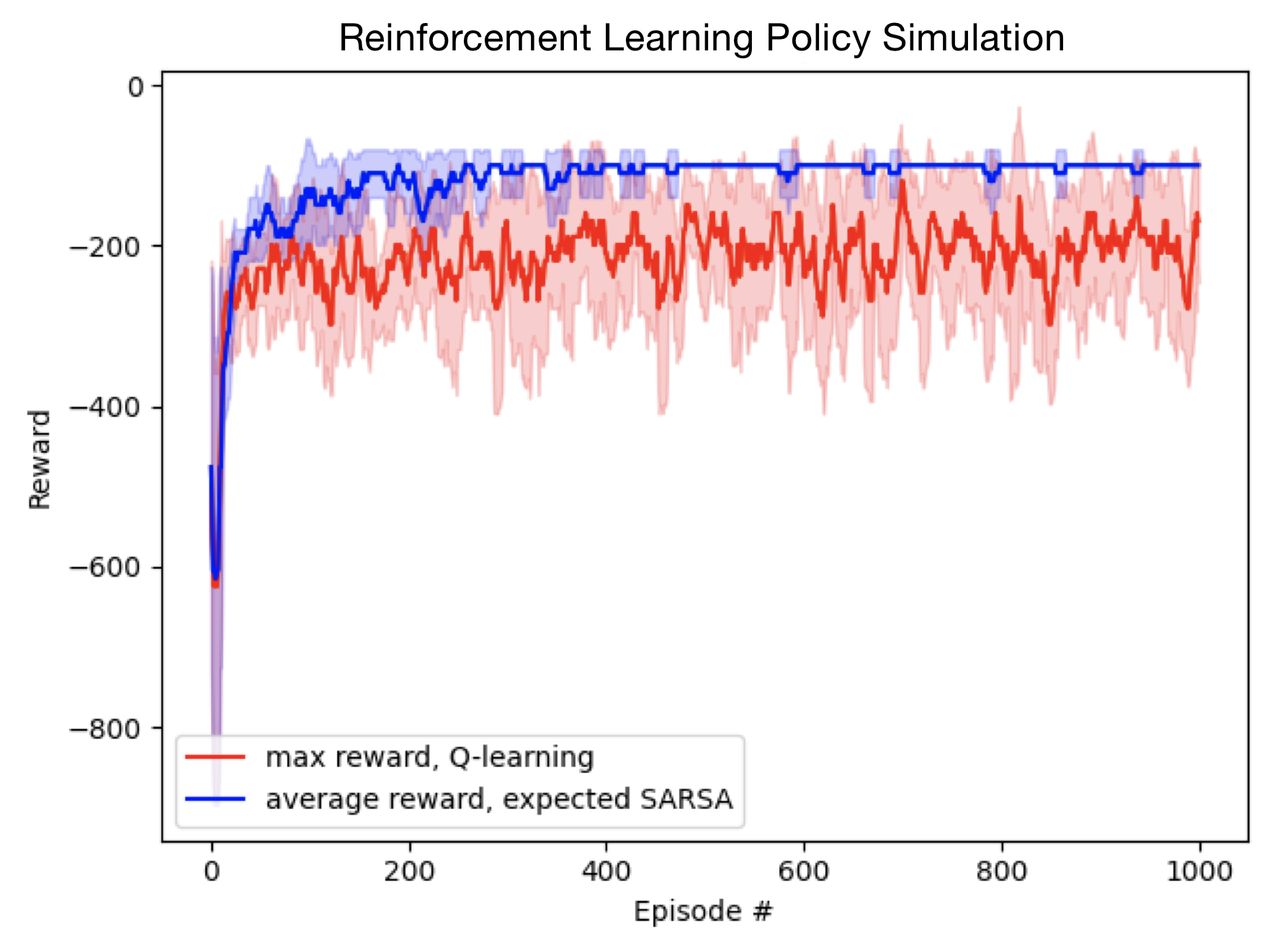}
  \caption{Simulation results showing the comparison between Expected SARSA and Q-learning. Expected SARSA was selected to deploy with the robot.}
  \label{fig:rlSimResults}
\end{figure}

Positional updates for the agent are contingent on the received signal of the olfaction sensor. 
The intensity of this signal translates to a positional update governed by basic kinematics. 
For each state transition with a weak olfaction signal (the robot is outside of the plume), the agent's reward is decremented by 5; for each transition with a strong signal (the robot is within the plume), the agent's reward is decremented by 1. 
For each agent, we use a discount factor of $\gamma=0.8$, a non-decaying learning rate of $\alpha=0.1$, and a value of $\epsilon=0.1$ that moderates the "greediness" of Expected SARSA.
Best results were achieved with a \textit{decaying} learning rate, but we found them difficult to repeat, so we keep the learning rate constant for results discussed here.
We train for 1000 episodes with 100 steps per episode for each task.
We cap the number of steps in order to induce a time limit on the robot.
Timing is critical in olfactory navigation because the target scent can equalize with the air and/or simply run out if it is not located in time.
Figure \ref{fig:rlSimResults} shows the simulation results.
After training, we save the RL model to employ it on the real robot.

\section{Experiments}
\begin{figure*}
  \centering
  \includegraphics[width=\textwidth]{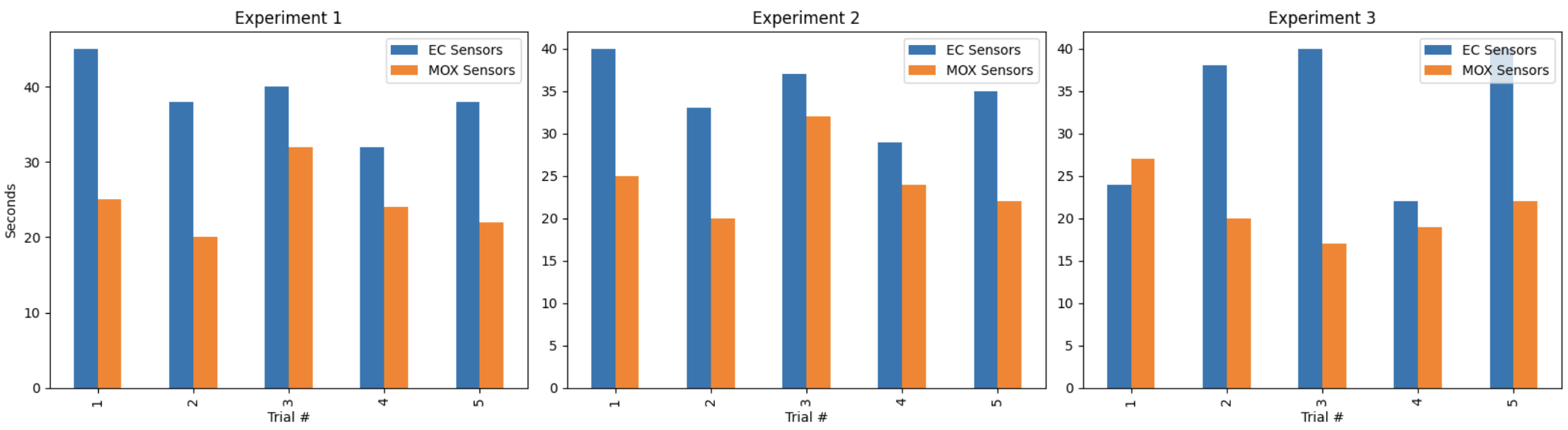}
  \caption{Results for Experiments 1, 2, and 3.}
  \label{fig:expResults}
\end{figure*}
We analyze three scenarios for each task: one scenario showing simple gradient following, one showing the RSSI-inspired belief maps algorithm, and a third showing OIO + RL.
For each experiment, we perform 5 trials and purge the room of the target compound before continuing on to the next trial.
Each trial ends when either the maximum time (60 seconds) has elapsed or the robot moves the sensor within 10 centimeters of the target compound.

In training, we strive to ensure that the robot starts "off plume" so that the robot knows it is not currently tracking the scent.
To facilitate this in reality, we must purge the room of the target compound to ensure that there is no trace of the scent to enable the robot to truly start "off plume".
We allow the target substance to be exposed to the surrounding air of the robot for 30 seconds before we begin tracking to allow a trackable plume to develop.
The target substance is held at the same location 2 meters away from the robot's shoulder; the robot is positioned such that it must at a minimum perform a 180-degree rotation to locate the odour.

Because we designed OIO to be sensor agnostic, there are some constraints that need to be established in order to make its methodology transferrable to different data types.
Each sensor type will have its own sensor noise which must be characterized experimentally in order to tune the Kalman filters.
Additionally, it is helpful to normalize the sampling time for each sensor type.
For experimental control, we hold velocity, acceleration, and jerk for all limbs on the robot constant at 10 $\deg/s$, 700 $\deg/s^2$, and 300 $\deg/sec^3$ respectively.

\subsection{Experiment 1: Gradient Following}
\vspace{-1mm}
For the simplest of our experiments, we analyze the robot's ability to simply ascend the gradient and maximize the observed RSSI value.
We note the average time to locate the source of the plume as 38.6 seconds and 24.6 seconds for electrochemical (EC) and metal oxide (MOX) sensors, respectively.
Both sensor types had similar standard deviations with 4.2 seconds and 4.1 seconds for EC and MOX sensors, respectively.


\subsection{Experiment 2: RSSI Belief Map Algorithm}
\vspace{-1mm}
For the RSSI-based algorithm, the number of steps required to find the odour source far exceeded 4 steps each time, which means several belief maps were constructed before a Voronoi vertex was found at a 4-sphere intersection.
The average time for the RSSI algorithm resulted in 36.4 seconds for the EC sensors and 24.4 seconds for the MOX sensors. 
While this is an improvement on the mean time to locate the odour source, the consistency between each trial decreased with standard deviations of 6.4 seconds and 4.3 seconds for EC and MOX sensors respectively.


\subsection{Experiment 3: Reinforcement Learning}
\vspace{-1mm}
The RL algorithm returned the highest volatility, but also returned the single fastest trial for both sensor types.
The average time to locate the odour source resulted in 32.8 seconds and 21.0 seconds for EC and MOX sensors respectively.
With these low average times also come higher volatility showing a standard deviation of 8.1 seconds for the EC trials and a standard deviation of 8.1 seconds for the MOX trials.

 
\section{Discussion}
\vspace{-1mm}
The results of each experiment show strengths and weaknesses of each algorithm.
There is evidence that RL does give some advantage in helping the robot more quickly localize the odour, but the extra volatility produced makes the algorithm undesirable for real applications where time consistency is desired to ensure a regular cadence of task completion.
Interestingly, the only trial where EC sensing achieved a lower time than MOX sensing was observed with the RL algorithm which gives promise that RL may be able to benefit the slow response times of EC sensor measurements.
More trials are needed in order to more appropriate quantify the exact benefit RL would provide in our olfactory pick-and-place scenarios.
We note that most success in modern visual-based pick-and-place tasks for robots is attributed to RL \cite{chi2024diffusion}, so we expect olfactory-based tasks to also follow in this manner.

The RSSI-inspired algorithm did show an improvement on regular gradient following, but not to the degree as the authors originally expected.
However, we take comfort in the fact that the near-identical completion times associated between both algorithms provides evidence to support that radio frequency principles can be applied to olfactory signals.

In the final analysis, the simplicity of simple gradient optimization seems to justify the slightly longer average odour localization times in comparison to the other two methods.
This gives nod to the success of vanilla OIO and the ability of kinematics to be fused with olfactory signals for compelling inertial odometry techniques.
With no baseline performance with which to compare our results, we note that our experiments undoubtedly leave room for future optimizations, but emphasize the fact that the robot successfully localized the odour source over all three tasks without timing out--a minimum threshold for our experimental success.

\section{Conclusion}
\vspace{-1mm}
In consideration of our results, we find evidence to support that navigation via scent can be accomplished by fusing olfaction with inertial odometry.
We propose a basic framework for OIO that provides compelling performance over multiple algorithms and sensors on a real robot, emphasizing OIO's generalization.
Our work here demonstrates how olfactory-oriented pick-and-place-style tasks can be  performed in the real world and establishes groundwork from which others can build and improve our results.

We note that our experiments give the robot no obstacles to avoid as it is localizing the odour source.
In future work, path planning improvements will be incorporated to allow the robot to use vision to avoid obstacles such as that proposed by Daescu and Malik in \cite{daescupath}.
Additionally, we are working to make hardware improvements that will encourage faster and more sensitive olfactory tracking techniques.
OIO shows great promise in following the footsteps of similar navigation techniques such as VIO and dead reckoning, and this is further enabled by RL.
We hope our work inspires more research in machine olfaction and expands awareness about the potential for scent-based navigation.

\section*{Acknowledgment}
\vspace{-1mm}
We wish to thank members of the Department of Computer Science for enabling access to their labs and resources for which to train our robots and validate our methodology.

\bibliographystyle{IEEEtran}
\bibliography{sample-base}





\end{document}